# U-shaped Transformer: Retain High Frequency Context in Time Series Analysis

Qingkui Chen[1], Yiqin Zhang[1]

[1] *School of Optical-Electrical and Computer Engineering, University of Shanghai for Science and Technology*
*Shanghai, China*

## Abstract

Time series prediction plays a crucial role in various industrial fields. In recent years, neural networks with a transformer backbone have achieved remarkable success in many domains, including computer vision and NLP. In time series analysis domain, some studies have suggested that even the simplest MLP networks outperform advanced transformer-based networks on time series forecast tasks. However, we believe these findings indicate there to be low-rank properties in time series sequences. In this paper, we consider the low-pass characteristics of transformers and try to incorporate the advantages of MLP. We adopt skip-layer connections inspired by Unet into traditional transformer backbone, thus preserving high-frequency context from input to output, namely **U-shaped Transformer**. We introduce patch merge and split operation to extract features with different scales and use larger datasets to fully make use of the transformer backbone. Our experiments demonstrate that the model performs at an advanced level across multiple datasets with relatively low cost.

## 1  Introduction

Long sequence time-series forecasting(LSTF) is widely applied in real life scenarios, e.g., Energy, transportation, meteorology, machinery[1]. In the highly popular field of deep learning in recent years, researchers primarily used recurrent neural networks (RNNs) and convolutional neural networks(CNNs) for processing time series data. However, extensive practice has revealed that RNNs have limited capabilities in effectively capturing and integrating long-range temporal information. CNNs do require stacking multiple layers to enlarge their receptive fields and may suffer from uneven receptive fields[2]. These approaches could also be inefficiency when handling long sequence time series.

The transformer architecture[3], introduced in 2017, has demonstrated impressive performance in domains such as natural language processing(NLP) and computer vision(CV). Its utilization of self-attention mechanisms enables the effective capture of long-range dependencies, efficiently extracting key information from the context. Researchers quickly realized the immense potential of this characteristic in time series analysis. As a result, a considerable number of transformer-based models specifically designed for time series data were proposed[4].

Recently, some researchers have revisited transformer-based time series prediction models[9] and highlighted that using simple Multilayer Perceptron networks(MLPs) can surpass the accuracy of transformer-based models on various datasets. Moreover, MLP networks have demonstrated massive advantages in terms of computational efficiency, achieving a decisive victory in this aspect. This research has sparked great impact by suggesting that transformers may not be an ideal foundation for constructing time series models. However, several studies[4], including ours, have maintained skepticism towards this viewpoint. Given that transformers have proven to be effective in multiple popular domains and have even found actual applications[5][6][7][8], the representational ability of transformers should be relatively strong. We believe the above research serves as guidance for transformer design rather than negation, and attempts to combine the two to achieve better results.

MLPs have relatively simple structures and limited fusion capabilities. If such a network works well on time series processing, it implies that certain information in the time series can be directly output as predictions without excessive processing. This brings to mind the renowned model Unet[12] in the field of image segmentation which does great on preserving high frequency context. While Unet is commonly used in CV domains and CNN-based, we considered that time series data also exhibit high and low-frequency characteristics, and this concept could be beneficial for time series analysis. Therefore, we propose U-shaped Transformer architecture with skip connection from encoder to decoder, allowing high-frequency data to quickly approach the head network without excessive processing. Our design transforms the residual connections of Unet's 2-D convolution into transformer residual connections for processing 1-D data

To enable the model to effectively operate across different temporal patterns, it is crucial to integrate information from different time scales. Thus, we introduced Learnable Patch Merge and Patch Split. They can be incorporated into each layer of the Transformer to assist the attention mechanism in focusing on information from different temporal scales more effectively.

In the field of CV and NLP, considering the effectiveness of transformers on large datasets, the combination of pretraining and fine-tuning has become a common approach[10][11]. As certain time series data can be highly valuable and difficult to obtain, such method could also serve as transfer learning which help to alleviate the shortage of rare time series data. Therefore, we designed patch-based pretraining tasks and fine-tuning tasks in our model training process. This approach not only improves model efficiency but also provides more possibilities for downstream tasks.

Our experiment results showing our model is highly accurate and efficient. The previous work actually degraded performance as the length of the input sequence increased. Results also showing our model no longer has such a problem. We also analyze multiple aspects that may influence the performance.

In summary, our contributions mainly consist of three aspects:

1. We propose the U-shaped transformer with skip connection, which allowing high-frequency data to quickly approach the output end of the neural network without excessive processing.
2. We introduce learnable patch merge and split, improving model's ability to extract features of different scales.
3. We designed patch-based pretraining tasks and fine-tuning tasks and introduce a larger dataset for pretraining to further exploit transformers' potential.

## 2 Related Work

**Unet**[12] utilizes cross-layer residual connections that allow information to flow from the bottom layers to the top layers, preserving high-frequency information. It has achieved impressive results in image segmentation and has recently made significant strides in areas such as image generation.[13]

**DLinear**[9] strongly criticized the effectiveness of transformers in time series prediction and demonstrated that simple MLP models can achieve SOTA performance on multiple datasets. Their work made us realize that in time series data, a considerable portion of the data can be represented as long-term predictions without requiring excessive processing.

**Vision Transformer**[14], **PatchTST**[15], and other patch-based[16][17] or token-based[18][19] models have introduced the idea of patch embedding, where they divide the input data into groups or patches within a certain range. Attention is then computed at patch level, allowing them to maintain prior knowledge of the input data and reduce the computational complexity of attention calculations. This approach proves beneficial for handling large-scale data and improving computational efficiency. LogTrans[21], FEDformer[22], Pyraformer[23], Autoformer[25], AnomalyBERT[24], Wang[26] and other works focus on improving attention mechanism and introducing decomposition to achieve higher accuracy with fewer computational costs. However, this approach breaks away from the traditional design of Transformers, and there are many implementation details to consider.

Zhang[20] reevaluated several key factors and underlying principles of transformers in time series analysis tasks and provided numerous theoretical analyses to support their findings. Their work demonstrated that transformers are fully capable of handling such tasks.

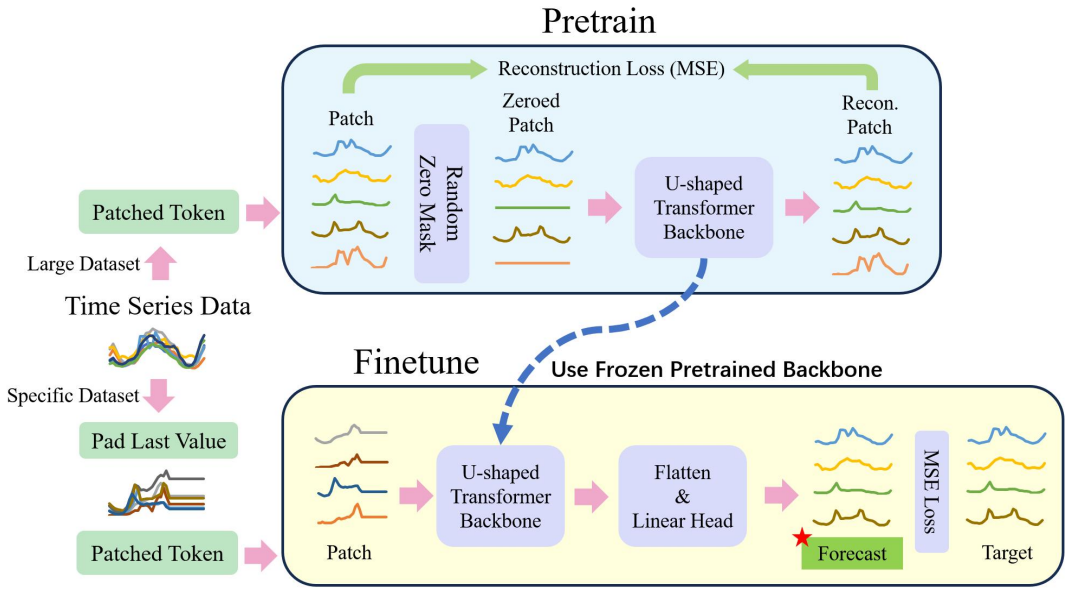

**Figure 1:** Process Overview. We employed the pretraining and fine-tuning approach to enhance the representational capacity of the model's backbone for time series data. Such a design can fully leverage the extraction capabilities of transformers on large datasets, making it highly efficient during fine-tuning. Additionally, it is easy to modify to adapt to different downstream tasks.

Informer[28], Wilhelm[29], and Trindade[30] introduced more time series dataset, covering multiple industries. These data are crucial to deep learning. Data used in our work tend to be without defect, they have gone through pre-process by data providers to reduce data anomalies.

Kiran[31] introduced Yformer which combining UNet design with Transformer. Their design requires two separate encoders to generate distinct high-dimensional feature representations for known and future sequences. To enhance the efficiency of the attention mechanism, they incorporated masks into the attention computation, allowing the model to extract information only from the known portions. YFormer also devised a more intricate loss function by adding an additional reconstruction loss, reinforcing the fitting ability during training.

## 3  Method

The overall process of our proposed time series prediction approach is illustrated in *Figure 1*. First, we employ the U-shaped Transformer backbone to reconstruct zeroed patches as pretrain tasks on large datasets. Second, we fine-tune the backbone-frozen pretrained model by training the head network for the designated task. In our work, our main focus is on time series prediction. Such design is easy to modify for other downstream tasks which only require different head designs.

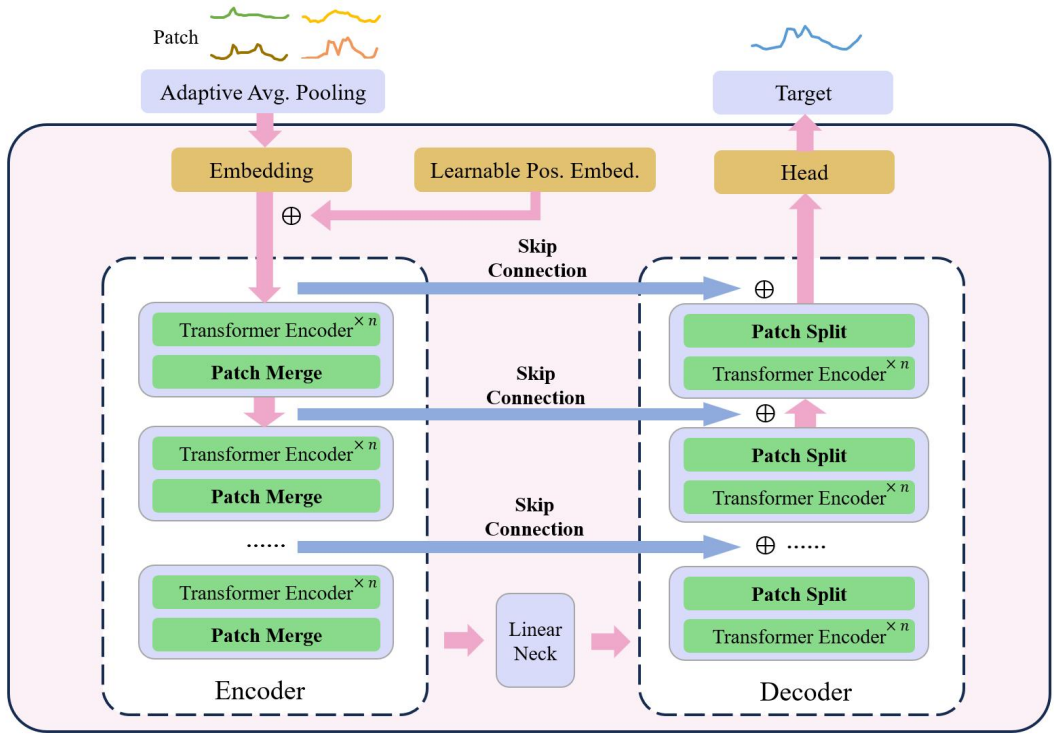

**Figure 2:** Network Model Overview. The data from each layer can directly traverse from the encoder to the decoder and is directly added together. Patch merge and split operations have been added between transformer groups. $n = 1$ in our original designs.

We define $L$ as length of input time series points, $T$ as length of target forecast time series points, $x_t$ as the time sample data at $t$ point. Thus, the data fed into the model equals to $X = \{x_1, x_2, \cdots, x_{L-1}, x_L\}$, forecast value is $\widehat{X} = \{\widehat{x_{L+1}}, \widehat{x_{L+2}}, \cdots, \widehat{x_{L+T-1}}, \widehat{x_{L+T}}\}$, the truth data corresponding to the forecast equals to $Y = \{x_{L+1}, x_{L+2}, \cdots, x_{L+T-1}, x_{L+T}\}$.

### 3.1 U-shaped Transformer

*Figure 2* is the overview of our backbone model which is based on stacking multiple transformer layers. Several transformer layers form a group, and after the group processing, patch merge or patch split operations are performed to integrate features from different scales. Multiple skip-layer connection serves as residuals from encoder to decoder.

**Skip-layer Connection.** As a kind of residual, skip-layer connection allows high-frequency data to quickly approach the output end of the neural network without excessive processing. The input data of $i_{th}$ transformer group $I_i^{enc}, i \in [1,2,\cdots,n_{layer}]$ are fed into the $i_{th}$ decoder output $O_i^{dec}, i \in [1,2,\cdots,n_{layer}]$ with same shape.*(Formula 1,2)* Former methods majorly use Concat and Linear Layer when handling data from skip connection. According to our experiments, MLPs used the operation above has little impact on accuracy. Therefore, we replace it with a simple summation of the corresponding dimensions.*(Formula 1)* During the encoding process, as the data progresses downward, high-frequency features are continuously filtered out while overall features are extracted. In the decoding process, the overall features are continuously restored with detailed information from skip-layer residual connection, ultimately resulting in a time series representation from the backbone that combines both high and low frequency features.

$$I_i^{dec} = Dec_{i+1}(I_{i+1}^{dec}) + O_i^{enc}, i \in [1, 2, \cdots, n_{layer}] \quad (1)$$

$$O^{dec} = O_1^{dec} + I_1^{enc} \quad (2)$$

Where $I$ implies input and $O$ implies output. $O^{dec}$ is the final output of decoder.

**Patch Merge and Patch Split.** Patch operations are critical components for the model to obtain feature at different scales, as they directly change what information will be included in the basic units(key-value pairs) of attention computation. Traditional approaches*(Figure 3a)* often segment time series into double arrays and treat them as independent channels.*(Formula 4)* We consider this approach to be coarse since the patch information from different channels at the same time step does not come from adjacent regions. Therefore, we instead use a convolution with a kernel size of 2 and a stride of 2 as patch merge, which doubling the number of channels.*(Figure 3b)* This ensures that prior is not fragmented, leading to better scale fusion effects. During the decoding process, we correspondingly employ transpose convolutions as Patch Split operation.

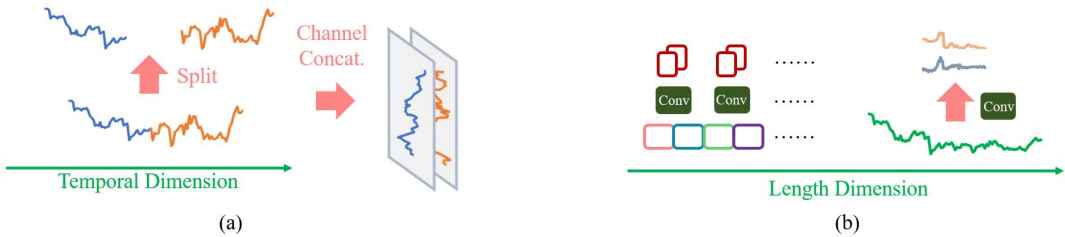

**Figure 3:** Patch Split and Merge. (a) is a simple patch merge operation using split and concat without any learnable parameters. (b) is our method, introducing learnable components should improve models' adaptive ability on patterns of different scales.

$$S_1^{\frac{p}{2} \times d} = \{x_1, x_2, \cdots, x_{\frac{p}{2}}\} \in X_i^{P \times D}, \quad S_2^{\frac{p}{2} \times d} = \{x_{\frac{p}{2}+1}, x_{\frac{p}{2}+2}, \cdots, x_p\} \in X_i^{P \times D} \tag{3}$$

$$X_o^{\frac{P}{2} \times 2D} = \begin{cases} C_1 = S_1^{\frac{p}{2} \times d} \\ C_2 = S_2^{\frac{p}{2} \times d} \end{cases} \tag{4}$$

Where $P$ is the number of input patch, $D$ is the input patch size, $C_i$ means i-th channel.

$$X_o^{\frac{P}{2} \times 2D} = \{Conv_1^2(x_i, x_{i+1}) | i \in \{1,3,5, \cdots, P-1\}\} \tag{5}$$

$$Conv_i^j(X) = bias(C_j) + \sum_{k=1}^{C_i} W(C_j, k) \times input(x_k \in X) \tag{6}$$

Where $Conv_i^j$ denotes the convolution will project channel from $i$ to $j$, $W$ is learnable parameter in convolution layer.

**Patch Embed.** Before dividing the time series into patches, our model utilizes an Adaptive Average Pooling layer to adjust the input sequence to the desired length for self-attention. Noted that during fine-tuning and forecast inference, the padded data appended after the input sequence is fixed (i.e., fixed $\frac{len(input)}{len(output)}$). Since the input length is adjustable, this can be considered as an approximation of a variable output time length. We employ pointwise convolution as the embedding method to map each patch to a higher-dimensional space. Afterwards, we prepare Learnable Relative Position Encodings for each patch, which are directly added to the embedded patches to enhance the accumulation of prior knowledge across patches.

### 3.2 Reconstruction Pretrain Task and Fine-tune

As has been mentioned above, we employ a patch-based embedding approach to divide the time series data into smaller blocks. We use patch reconstruction task similar to PatchTST[15] as our pretrain task. We believe that reconstructing the unmasked patches can improve the model's robustness to zero-valued noisy data. After pretraining, we perform fine-tuning on task-specific head, which is responsible for generating the task objectives. We freeze all components except head networks at this stage. The loss function of our method is described by *Formula 7,8*.

$$L_{pretrain} = MSE \begin{cases} Source = Patch(X) \\ Target = F_{backbone}[ZeroMask(Source)] \end{cases} \tag{7}$$

$$L_{finetune} = MSE \begin{cases} Source = \hat{X} = Head\{F_{backbone}[Patch(X)]\} \\ Target = Y \end{cases} \tag{8}$$

### 3.3 Data Preprocessing

To facilitate better model generalization and exploit transformer's potential on large datasets, we employed a larger dataset in our experiments, including Electricity Transformer Temperature (ETT)[28], traffic, electricity, weather, ILI, Dhaka Stock Exchange, Earth Surface Temperature, Electric Motor Temperature, Sin function, and Log function.

**Look Back Window.** The datasets above exhibit significant differences in their frequency domain characteristics, as well as considerable variations in their lengths. To address this problem, we employed a Variable-length Jittered Sliding Window *(Figure 4, Formula 9)* approach for data augmentation. The length of the interval directly controlled the number of generated samples, helping alleviate the issue of inconsistent lengths among multiple datasets. This data augmentation method enabled the model to extract features from the dataset effectively across different epochs of training.

$$\begin{cases} Sample_\ell = \{x_{i \times \tau}, x_{i \times \tau+1}, \cdots, x_{i \times \tau+L-1}, x_{i \times \tau+L}\} \in X \\ i = \ell + \varepsilon, \ \varepsilon \in \left[0, \frac{\tau}{2}\right], \ \ell \in \left[0, \frac{len(X)}{\tau} - L - \frac{\tau}{2}\right] \end{cases} \tag{9}$$

Where $\tau$ is sampling stride, $\ell$ is original sample index.

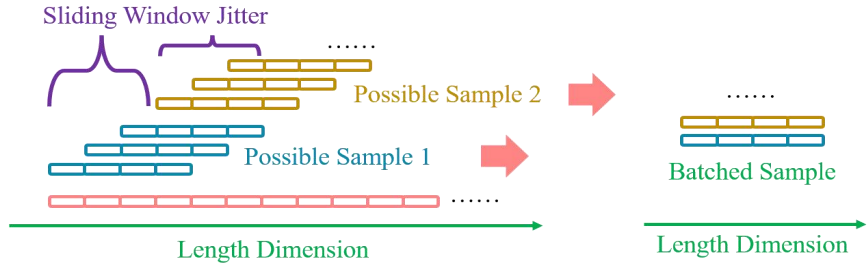

**Figure 4:** Jittered Sliding Window Augmentation. This example is based on Sample Length 2 and Stride 3. Stride and Jitter range could be used to control the intensity of data augmentation. Such design brings quick and full-scale data fetch.

**Weighted Sampling.** To further mitigate data imbalance, we employed weighted random sampling, where the number of samples taken from different datasets during training is balanced. The weights for sampling point $x$ in dataset $d$ were $W_x = \frac{1}{X_d} \cdot \frac{1}{D}$, where $X_d$ is number of samples in dataset $d$, $D$ is the number of datasets used during training.

**Normalization.** In order to ease the issue of training instability in transformer-based networks, we normalize each mini-batch to follow a standard normal distribution[32]. Based on our experiments, omitting this step significantly increases the possibility of gradient explosion. Noted that DLinear[9] and other works(e.g. [21][22][23][25]) normalize dataset based on the mean and variance of each channel in the entire dataset, while PatchTST[14] introduced Instance Normalization[32] which performs normalize based on the sample's mean and variance. *(Formula 11)* Such different design makes it unreasonable to directly compare MSE losses between them because of their different normalization methods. As a result, we unify our data preprocessing method following *Formula 12* and reproduce DLinear and PatchTST in the following chapter.

$$S_i = \{x_i, x_{i+1}, x_{i+2}, \cdots, x_{i+L-1}, x_{i+L}\} \in X, \ 0 \le i \le N_X - L \tag{10}$$

$$S_i^{DLinear} = \frac{S_i - \mu(X)}{\sigma(X)} \qquad S_i^{PatchTST} = \frac{S_i - \mu(S_i)}{\sigma(S_i)} \tag{11}$$

$$S_i^{Ours} = \begin{cases} \frac{S_i - \mu(S_i)}{\sigma(S_i)}, & \sigma(S_i) \ge 0.01 \\ S_i - \mu(S_i), & 0 \le \sigma(S_i) \le 0.01 \end{cases} \tag{12}$$

In summary, our data preprocessing method ensures efficient extraction of features from different parts of the dataset and effectively alleviates the issue of data imbalance during joint dataset training.

## 4 Experiments

In our experiment, the model is trained and evaluated following the process shown in *Figure 1*. We primarily conducted experiments on patch size ∈ [16, 128] and patch stride ∈ [16, 512]. We design two configurations named small and base, their detailed setting is shown in *Table 1*.

**Table 1:** Key Setting of Each Model.

| Model | $L$ | $T$ | Patch Size | Patch Stride | Number of Patches |
|---|---|---|---|---|---|
| DLinear | 336 | 720 | N/A | | |
| PatchTST | 336 | 720 | 16 | 8 | 131 |
| Ours - small | 512 | 1024 | 32 | 32 | 48 |
| Ours - base | 3072 | 1024 | 32 | 32 | 128 |

### 4.1 Pretrain

As mentioned above, we aimed to enhance our model's performance on specific tasks by extensively pretraining it on larger datasets. Therefore, we utilized a total of 13 datasets(ETTm1, ETTm2, ETTh1, ETTh2, electricity, national illness, traffic, weather, Dhaka Stock Exchange, Earth Surface Temperature, Electric Motor Temperature, Sin, and Log) for pretraining.

Our results*(Figure 5)* of the pretrain loss show that the performance improves as the number of U-shape backbone layers increases. Note that not only does the performance improve within the same epoch, but it also improves within the same computational cost. Increasing the number of layers enhances the model's ability to extract features from data at different scales, further confirming the effectiveness of Patch Merge and Patch Split operations in improving the model's representational capacity.

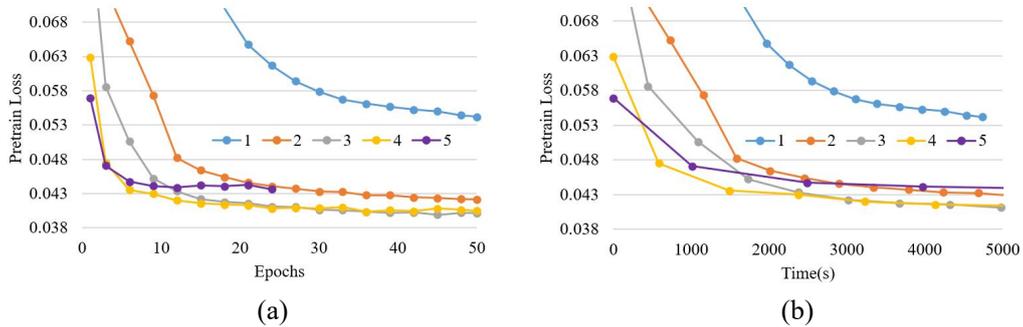

**Figure 5:** Pretrain Performance under Different Number of Layers. (a) compares the performance of the models at the same epoch and (b) is at the same time cost. More layers bring better accuracy and speed. Only a few epochs are needed to fit.

The attention weight heat map is illustrated in *Figure 6*. In the encoder, the attention allocation of the lower layers appears relatively uniform (please note the scale on the right-hand side). This allocation also exhibits a clear pattern: each query tends to attend to keys at fixed interval sizes, resulting in regular diagonal stripes in the heat map indicating that the lower-level attention has captured the frequency characteristics of the data. As we progress deeper into the layers, the number of key-value pairs gradually decreases due to the Patch Merge operation, causing the heat map to become coarser. However, it still maintains two distinctive features:

1) There is a significant increase in attention near the query: represented by a bright diagonal line in the heat map.
2) The attention weights exhibit interval patterns across different scales: represented by parallel bright interval segments in the heat map.

We give more discussion in **A.4**

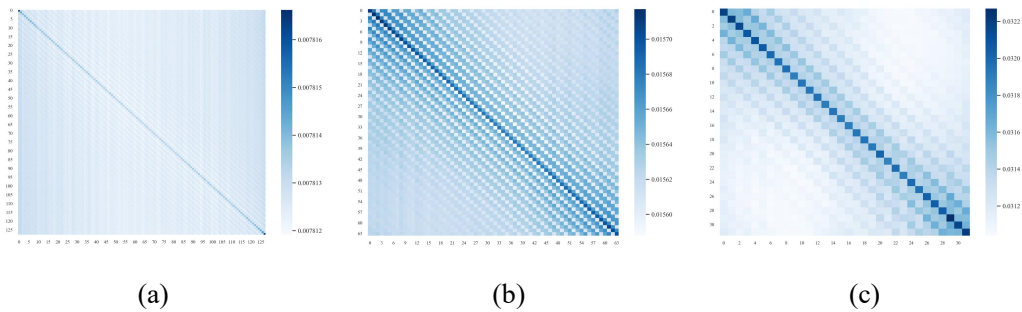

(a)           (b)           (c)

**Figure 6:** Encoder Attention Map. Figure a, b and c respectively visualize the attention score heat maps in the 1st, 2nd and 3rd layers. The regular interval patterns in the attention indicate that the model considers frequency features when extracting information from the time series. Adjacent tokens receive more attention, with this allocation being relatively mild in the initial layer. As the model goes deeper, the differences in allocation gradually increase.

## 4.2 Finetune

*Table 2* shows the major results of our finetuned models' performance on time series forecasting tasks, which is a crucial evaluation factor. The results demonstrate that as the input sequence length increases, the model's prediction performance consistently improves and eventually surpasses that of DLinear. Due to hardware and resource constraints, we did not reproduce more mainstream Transformer-based models. However, based on multiple research findings, DLinear's performance exceeds that of the aforementioned models.[4][33][34][35]

The pretraining and fine-tuning approach significantly reduces computational resources. *Figure 8* illustrates the fine-tuning efficiency of different models. Just one epoch of training is sufficient to fine-tune the model to a considerable level, surpassing the performance of DLinear within the same computational time. This demonstrates the speed and effectiveness of our method. *Figure 9* shows several inference examples using our model, where our models' outputs have obvious low and high frequency patterns.

**Table 2:** Loss on Test Dataset. We evaluate the loss between $\{\hat{x}_{L+1}, \hat{x}_{L+2}, \cdots, \hat{x}_{L+T-1}, \hat{x}_{L+T}\} \in \hat{X}$ and $\{y_{L+1}, y_{L+1}, \cdots, y_{L+T-1}, y_{L+T}\} \in Y$, where $T \in \{96, 192, 336, 720\}$, $L_{DLinear} = 336$.

| Model | Step | Overall* MSE | Overall* MAE | ETTm1 MSE | ETTm1 MAE | traffic MSE | traffic MAE | weather MSE | weather MAE | electricity MSE | electricity MAE | Earth Surface Temperature MSE | Earth Surface Temperature MAE | Dhaka Stock Exchange MSE | Dhaka Stock Exchange MAE |
|---|---|---|---|---|---|---|---|---|---|---|---|---|---|---|---|
| DLinear | 96 | 0.499 | 0.470 | 0.607 | 0.579 | 0.323 | 0.310 | 0.675 | 0.596 | 0.225 | 0.320 | 0.255 | 0.313 | **0.798** | 0.624 |
| | 192 | 0.525 | 0.503 | 0.616 | 0.590 | **0.352** | 0.337 | 0.691 | 0.607 | 0.245 | 0.342 | 0.343 | 0.376 | **0.828** | 0.644 |
| | 336 | 0.507 | 0.492 | 0.637 | 0.608 | 0.400 | 0.356 | 0.649 | 0.583 | 0.258 | 0.349 | 0.432 | 0.449 | **0.803** | 0.630 |
| | 720 | 1.250 | 0.732 | 1.005 | 0.716 | 0.662 | 0.476 | 1.467 | 0.808 | 0.660 | 0.483 | 0.921 | 0.659 | 1.898 | 0.874 |
| PatchTST | 96 | 0.616 | 0.472 | 0.916 | 0.702 | 0.327 | **0.289** | 1.112 | 0.741 | NaN▲ | | **0.166** | **0.242** | NaN▲ | |
| | 192 | 0.803 | 0.566 | 1.049 | 0.756 | 0.381 | **0.325** | 1.535 | 0.894 | | | **0.252** | **0.297** | | |
| | 336 | 0.971 | 0.646 | 1.237 | 0.837 | 0.458 | 0.369 | 1.820 | 0.995 | | | 0.384 | **0.362** | | |
| | 720 | 1.914 | 0.893 | 1.823 | 1.007 | 0.736 | 0.491 | 2.829 | 1.218 | | | 0.994 | 0.582 | | |
| Ours - small | 96 | 0.453 | 0.437 | 0.566 | 0.552 | **0.312** | 0.306 | 0.623 | 0.561 | 0.214 | 0.314 | 0.233 | 0.305 | 0.798 | 0.570 |
| | 192 | 0.515 | 0.481 | 0.619 | 0.585 | 0.353 | 0.331 | 0.691 | 0.605 | 0.241 | 0.340 | 0.322 | 0.360 | 0.869 | 0.630 |
| | 336 | 0.527 | 0.508 | 0.628 | 0.599 | **0.396** | **0.344** | 0.687 | 0.597 | **0.243** | 0.348 | 0.404 | 0.438 | 0.906 | 0.618 |
| | 720 | 0.561 | 0.518 | 0.676 | 0.622 | **0.582** | **0.459** | 0.722 | 0.608 | 0.349 | 0.409 | 0.607 | 0.586 | **0.841** | 0.624 |
| Ours - base | 96 | **0.366** | **0.376** | **0.519** | **0.516** | 0.333 | 0.330 | **0.506** | **0.479** | **0.186** | **0.286** | 0.212 | 0.285 | 0.851 | **0.505** |
| | 192 | **0.411** | **0.411** | **0.537** | **0.522** | 0.442 | 0.436 | **0.554** | **0.499** | **0.208** | **0.301** | 0.298 | 0.347 | 0.939 | **0.553** |
| | 336 | **0.433** | **0.423** | **0.576** | **0.553** | 0.435 | 0.405 | **0.598** | **0.525** | 0.259 | 0.351 | 0.375 | 0.419 | 0.971 | **0.550** |
| | 720 | **0.506** | **0.446** | **0.636** | **0.575** | 0.639 | 0.494 | **0.638** | **0.548** | **0.352** | **0.390** | 0.555 | 0.524 | 0.948 | **0.561** |

\* Mean performance across all pretrain dataset
▲ Failed for three tries of training. Model inference output contains Not a Number(NaN) value

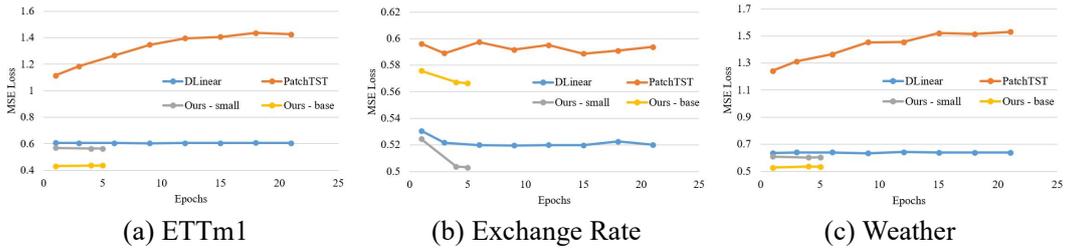

(a) ETTm1      (b) Exchange Rate      (c) Weather

**Figure 8:** Loss on Validate Set during Finetune. Our model only needs very a few of epochs to perform finetune. Besides, we also achieve excellent accuracy, surpassing DLinear in multiple datasets. In most cases, our tiny version model is enough to generate accurate forecast.

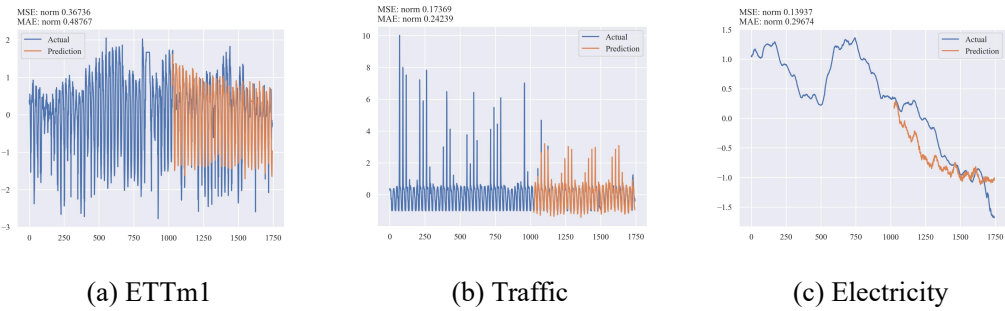

(a) ETTm1      (b) Traffic      (c) Electricity

**Figure 9:** Forecast Curve. The non-overlapping portion between the blue and yellow curves represents the input to the neural network, while the yellow curve represents the extrapolated values predicted by the model. Our models work well on datasets with different patterns, indicating the pretrain-finetune process is effective on various time series data.

# 5  Discussions

Our model not only achieves better results than DLinear under the same number of fine-tuning epochs but also performs better in terms of computational efficiency within the same time cost. For more comparisons regarding computational efficiency, please refer to **A.1**. We also analyze attention allocation between tokens with and without prediction padding in **A.4**.

We have observed that the performance of our small model does not always surpass the base one. Adjusting the length of the input sequence can contribute to improved accuracy for specific datasets. This might be because we adopt fixed-length patch generation, and when there are features coupled with this length in the sequence, individual patches can compute attention more effectively. We provide more experimental result in **A.2**.

We noticed a minimum size of patch is needed for our model to perform effectively. Different patch size directly leads to models' attention on different pattern scale of time sequence, thus influencing the performance. A larger patch size allows the model to extract long-range information more quickly, thereby reducing computational costs. However, excessively aggressive settings can have a severe impact on performance. We conduct ablation experiments in **A.3**.

# 6  Conclusions

We proposed a U-shaped Transformer with skip-layer residual connections that allows information to be transmitted to layers closer to the head network without excessive processing. This design preserves high-frequency features in sequential data and compensates for the low-pass characteristics of Transformers. The patch-based design approach and merge and split operations makes the network more versatile, and the combination of pretraining and fine-tuning significantly reduces redundant training costs. Our model is also easy to modify and can be applicable to various time series analysis tasks.

# Appendix

## A.1 Efficiency

We analyzed the model's performance during the fine-tuning process. The result*(Figure A.1.1)* shows our model significantly reduced the computational cost and maintains superior accuracy. This demonstrates that the pretraining process effectively establishes the model's ability to extract meaningful features from sequential data.

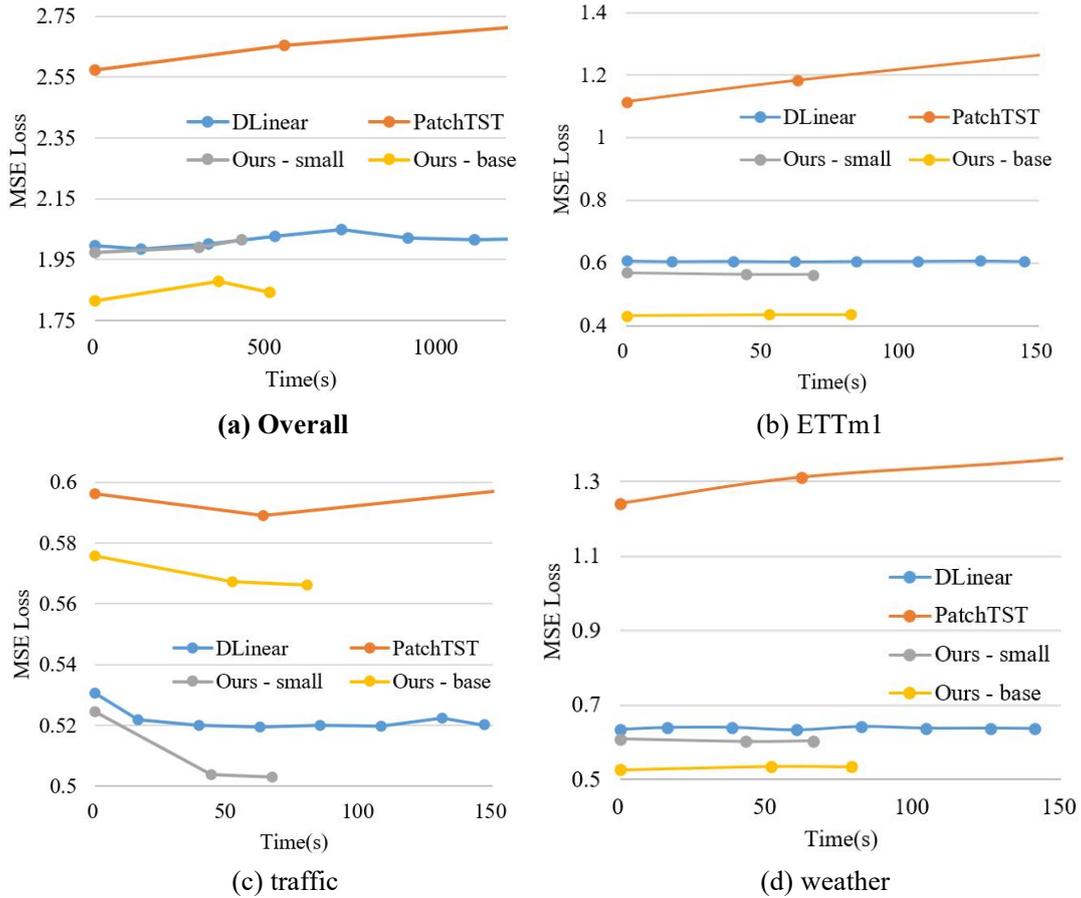

**Figure A.1.1:** Training Efficiency Comparison. Our pretrained models require only a short fine-tuning time to achieve near-optimal performance. The performance characteristics of our model closely resemble those of the efficient DLinear model.

The starting point of the experimental results curve mentioned above is at the end of the first finetune epoch. This approach is taken to avoid errors caused by the time-consuming initialization processes. In DLinear, the performance bottleneck tends to be the CPU and RAM rather than the GPU. This is because the network is relatively simple, to the extent that the GPU completes the neural network forward and backward propagation in a very short time, leading to the GPU mostly waiting for data. We use a 12700F CPU and made full use of multiprocessing for data preprocessing. However, the data transfer bottleneck could not be completely alleviated. Meanwhile, our model's performance bottleneck is still GPUs.

## A.2 Input Length

In some previous models, increasing the length of the lookback window actually led to a decrease in performance. This phenomenon could be attributed to the model being overwhelmed by an excessive amount of data. However, the DLinear model, with fewer learnable parameters, is less prone to getting trapped in deep confusion during the training process. In our model, the presence of skip-layer connections allows information to bypass extensive processing and be directly transmitted to the near output end. This kind of residual connection structure might help mitigate the performance degradation caused by deep confusion in the model. *(Figure A.2.1)*.

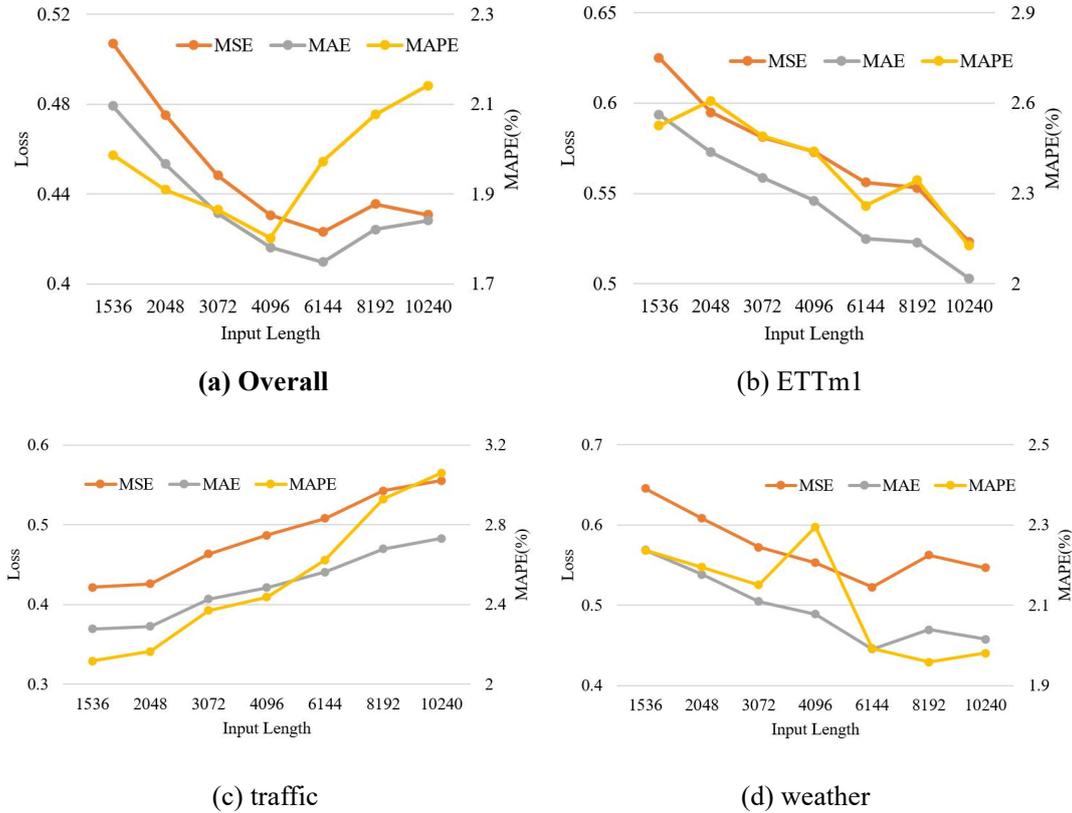

(a) Overall  (b) ETTm1

(c) traffic  (d) weather

**Figure A.2.1:** Performance Comparison under different Lookback Window Size. Please note that the values on the horizontal axis equals to $L + T$, while $T = 1024$. (a) is the mean performance across all dataset used for pretrain. Models' performance tends to increase as the window size become larger.

## A.3 Patch Size and Patch Stride

Different patch size, stride and layer numbers could directly lead to models' fusion ability on time sequences of different scales. *Table A.3.1* is under the patch size and stride fixed to 32. The number of key-value pairs will decrease from 128 to 4 when number of layers is set to 5. We discover that the performance improves as layer goes deeper, but deteriorate when it was too deep. *Table A.3.2* also shows the similar characteristics when modifying patch size.

**Table A.3.1:** Ablation between different Transformer Layer Settings

| Dataset | Overall | | | Dhaka Stock Exchange | | | Earth Surface Temp. | | | Electric Motor Temp. | | | ETTh1 | | |
|---|---|---|---|---|---|---|---|---|---|---|---|---|---|---|---|
| Layer | MSE | MAE | MAPE | MSE | MAE | MAPE | MSE | MAE | MAPE | MSE | MAE | MAPE | MSE | MAE | MAPE |
| 1 | 0.433 | 0.417 | 1.800 | **0.861** | **0.525** | **2.533** | 0.378 | 0.422 | 1.717 | 0.375 | 0.380 | 1.683 | **0.469** | **0.465** | 1.931 |
| 2 | **0.432** | 0.419 | 1.811 | 0.869 | 0.528 | 2.619 | **0.370** | **0.408** | **1.649** | 0.368 | 0.377 | 1.731 | 0.475 | 0.469 | 1.921 |
| 3 | 0.432 | **0.416** | **1.799** | 0.875 | 0.527 | 2.583 | 0.371 | 0.409 | 1.660 | **0.365** | **0.372** | 1.678 | 0.474 | 0.468 | **1.877** |
| 4 | 0.456 | 0.441 | 1.937 | 0.924 | 0.568 | 2.993 | 0.387 | 0.419 | 1.761 | 0.374 | 0.389 | 1.974 | 0.498 | 0.491 | 2.039 |
| 5 | | NaN▲ | | | NaN▲ | | | NaN▲ | | 1.141 | 0.765 | 4.681 | 0.494 | 0.488 | 2.031 |

▲ Failed at least once during three tries of training. Model inference output contains Not a Number(NaN) value

**Table A.3.2:** Ablation between Different Patch Size

| Dataset | Overall | | | Dhaka Stock Exchange | | | Earth Surface Temp. | | | Electric Motor Temp. | | | ETTh1 | | |
|---|---|---|---|---|---|---|---|---|---|---|---|---|---|---|---|
| Patch Size | MSE | MAE | MAPE | MSE | MAE | MAPE | MSE | MAE | MAPE | MSE | MAE | MAPE | MSE | MAE | MAPE |
| 8 | 0.452 | 0.443 | 1.935 | **0.897** | **0.557** | **2.899** | 0.403 | 0.444 | 1.853 | 0.381 | 0.406 | 1.797 | 0.502 | 0.494 | 2.026 |
| 16 | 0.453 | 0.445 | 1.907 | 0.915 | 0.565 | 2.941 | 0.402 | 0.439 | 1.821 | **0.371** | 0.39 | 1.91 | 0.5 | 0.492 | 2.029 |
| 32 | **0.45** | **0.435** | **1.884** | 0.926 | 0.566 | 2.941 | **0.391** | **0.432** | **1.789** | 0.373 | **0.389** | **1.766** | 0.492 | **0.487** | **1.995** |
| 64 | 0.476 | 0.454 | 2.06 | 0.952 | 0.596 | 3.351 | 0.442 | 0.46 | 1.985 | 0.397 | 0.411 | 2.112 | 0.528 | 0.515 | 2.127 |
| 128 | | NaN▲ | | | NaN▲ | | | NaN▲ | | | NaN▲ | | | NaN▲ | |

▲ Failed at least once during three tries of training. Model inference output contains Not a Number(NaN) value

Based on the discovery above, we believe modify neither layer nor patch size themselves could not achieve stable better results, so we further explore different configurations with same number of key-value pairs in the deepest layer.*(Table A.3.3)* All models performed well this time without any NaN output. Their accuracy is also similar, while larger patch size initially has wider extraction window and fewer layers reduced data fusion of different scales, thus canceling each other out.

**Table A.3.3:** Ablation under the same number of key-value pairs(16 in this table) in the deepest layer. P is patch size and L is num of layers. $L = 3072, T = 1024$.

| Dataset | Overall | | | Dhaka Stock Exchange | | | Earth Surface Temp. | | | Electric Motor Temp. | | | ETTh1 | | |
|---|---|---|---|---|---|---|---|---|---|---|---|---|---|---|---|
| Config | MSE | MAE | MAPE | MSE | MAE | MAPE | MSE | MAE | MAPE | MSE | MAE | MAPE | MSE | MAE | MAPE |
| P8 L5 | 0.449 | 0.439 | 1.911 | 0.875 | 0.535 | **2.541** | 0.379 | 0.415 | 1.689 | 0.384 | 0.402 | 1.766 | 0.479 | 0.474 | 1.967 |
| P16 L4 | 0.460 | 0.450 | 1.947 | 0.882 | 0.540 | 2.651 | 0.394 | 0.445 | 1.747 | 0.375 | 0.390 | **1.702** | 0.484 | 0.478 | 1.966 |
| P32 L3 | 0.451 | 0.434 | 1.952 | 0.874 | 0.529 | 2.561 | 0.375 | 0.408 | 1.646 | 0.368 | 0.379 | 2.005 | 0.475 | 0.469 | **1.916** |
| P64 L2 | **0.429** | **0.414** | **1.786** | **0.867** | **0.528** | 2.614 | **0.365** | 0.410 | 1.626 | **0.366** | **0.373** | 1.787 | **0.471** | **0.468** | 1.949 |
| P128 L1 | 0.451 | 0.425 | 1.876 | 0.870 | 0.534 | 2.660 | 0.369 | **0.408** | **1.599** | 0.379 | 0.384 | 2.213 | 0.511 | 0.482 | 1.943 |

## A.4 Attention Weight

Self-attention can establish connections over global distances with $O(1)$ complexity. We visualized the attention computation results of the first layer in each transformer block, which represents the average results over a single batch. The attention weight maps provide insights into the network's behavior in extracting information from different parts of the sequence. *Figure A.4.1* shows the attention weights for each layer of the Encoder. In each layer, the queries typically have higher attention towards the surrounding keys*(Figure A.4.2)*, which aligns with the prior assumption that sequences exhibit temporal continuity in the time dimension.

We have also observed an interesting phenomenon: queries may exhibit higher attention towards two adjacent keys and, in the next layer, once again prioritize adjacent two keys. These consecutive operations temporarily separate the information and then bring it back to the original

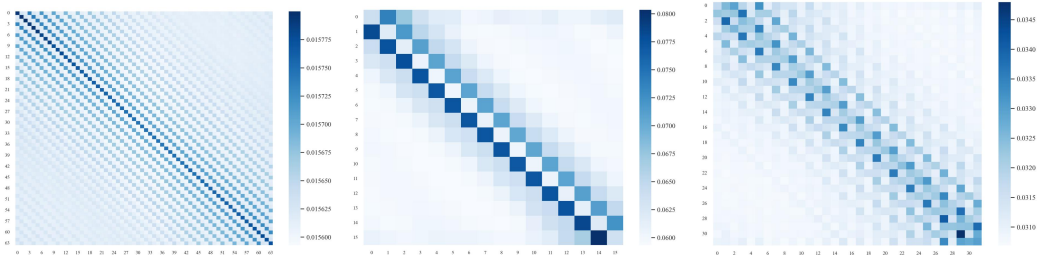

**Figure A.4.1:** Encoder Attention Weight in Pretrain.

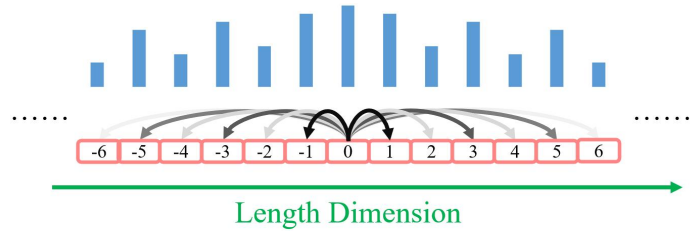

Length Dimension

**Figure A.4.2:** Attention Allocation Diagram with Interval Features. The model's attention allocation is essentially a frequency feature extraction, which enables each attention to focus on temporal features at a certain interval scale. Such features are prevalent across different layers.

position after combining it with other information. This behavior is somewhat similar to the concept in Swin-based models, although we did not explicitly introduce this idea in our model. We believe that this behavior, learned by the model itself, allows for better fusion capabilities between different key-value pairs.

During the non-pretraining process*(A.4.3)*, we add duplicate values of the last element to the end of the input sequence, causing attention to exhibit similar weight distribution among the later key-value pairs. However, due to the presence of learnable relative position encoding, there are subtle differences in this distribution. Experimental results indicate that our model exhibits significantly higher self-attention towards the information in real sequences, while the key-value pairs containing the portion to be predicted tend to pay more attention to the real sequence. In other words, attention is fusing information from known data sequences and forming future key-value pairs.

In *Figure A.4.3*, Red box indicating known values fusing themselves, showing strong self-attention pattern. Purple and yellow boxes show relative low attention for these parts of keys are unknown tokens which containing few information. Green box shows slight pattern indicting models are extracting specific context from known key-value pairs. Note that the attention weights of the decoder exhibit stronger attention biases, indicating that the model has focused on specific parts of the sequence. Particularly, the yellow box representing self-attention over the unknown sequence also shows some diagonal stripes. This suggests that the model has generated effective and consistent information for predicting the key-value pairs, rather than diffused extraction.

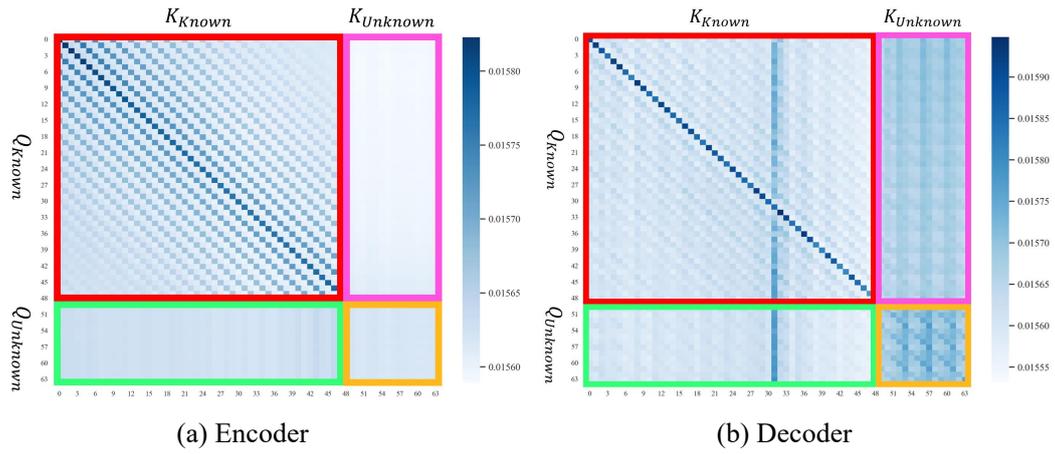

(a) Encoder  (b) Decoder

**Figure A.4.3:** Attention Weight in Forecasting.